# When do Numbers Really Matter?

Hei Chan and Adnan Darwiche
Computer Science Department
University of California
Los Angeles, CA 90095
{hei,darwiche}@cs.ucla.edu

## Abstract

Common wisdom has it that small distinctions in the probabilities quantifying a belief network do not matter much for the results of probabilistic queries. Yet, one can develop realistic scenarios under which small variations in network probabilities can lead to significant changes in computed queries. A pending theoretical question is then to analytically characterize parameter changes that do or do not matter. In this paper, we study the sensitivity of probabilistic queries to changes in network parameters and prove some tight bounds on the impact that such parameters can have on queries. Our analytical results pinpoint some interesting situations under which parameter changes do or do not matter. These results are important for knowledge engineers as they help them identify influential network parameters. They also help explain some of the previous experimental results and observations with regards to network robustness against parameter changes.

## 1 Introduction

Automated reasoning systems based on belief networks [9, 6] have become quite popular recently as they have enjoyed much success in a number of real-world applications. Central to the development of such systems is the construction of a belief network that faithfully represents the domain of interest. Although the automatic synthesis of belief networks—based on design information in certain applications and based on learning techniques in others—have been drawing a lot of attention recently, mainstream methods for constructing such networks continue to be based on traditional knowledge engineering (KE) sessions involving domain experts. One of the central issues that arise in such KE sessions is to assess the impact that changes to network parameters may have on queries of interest.

We have recently developed a sensitivity analysis tool, called SAMIAM (Sensitivity Analysis, Modeling, Inference And More), which allows domain experts to fine tune network parameters in order to enforce constraints on the results of certain queries. For example, it may turn out that $Pr(y \mid \mathbf{e})/Pr(z \mid \mathbf{e}) = 2$ with respect to a given belief network, while the domain expert believes that the ratio $Pr(y \mid \mathbf{e})/Pr(z \mid \mathbf{e})$ should be equal to 3. SAMIAM will then automatically decide whether a given parameter is relevant to this constraint, and if it is, will compute the minimum amount of change to that parameter which is needed to enforce the constraint.[1]

As we experimented with SAMIAM, we ran into scenarios that we found to be quite surprising. Specifically, there were many occasions in which queries would be quite sensitive to very small variations in certain network parameters.

**Example 1.1** *Consider Figure 1, which depicts a screen shot of SAMIAM. The depicted network has six binary variables, each of which has the values true and false. There are two query variables of interest here, fire and tampering. Suppose that the evidence $\mathbf{e}$ is report, $\overline{smoke}$: people are reported to be evacuating a building (in response to an alarm), but there is no evidence for any smoke. This evidence should make tampering more likely than fire, and the given belief network did indeed reflect this since $Pr(tampering \mid \mathbf{e}) = .50$ and $Pr(fire \mid \mathbf{e}) = .03$ in this case. We wanted,*

---
[1]We use the following standard notation: variables are denoted by upper-case letters ($A$) and their values by lower-case letters ($a$). Sets of variables are denoted by bold-face upper-case letters ($\mathbf{A}$) and their instantiations are denoted by bold-face lower-case letters ($\mathbf{a}$). For a variable $A$ with values *true* and *false*, we use $a$ to denote $A = true$ and $\overline{a}$ to denote $A = false$. Finally, for a variable $X$ with parents $\mathbf{U}$, we use $\theta_{x|\mathbf{u}}$ to denote the network parameter corresponding to $Pr(x \mid \mathbf{u})$.

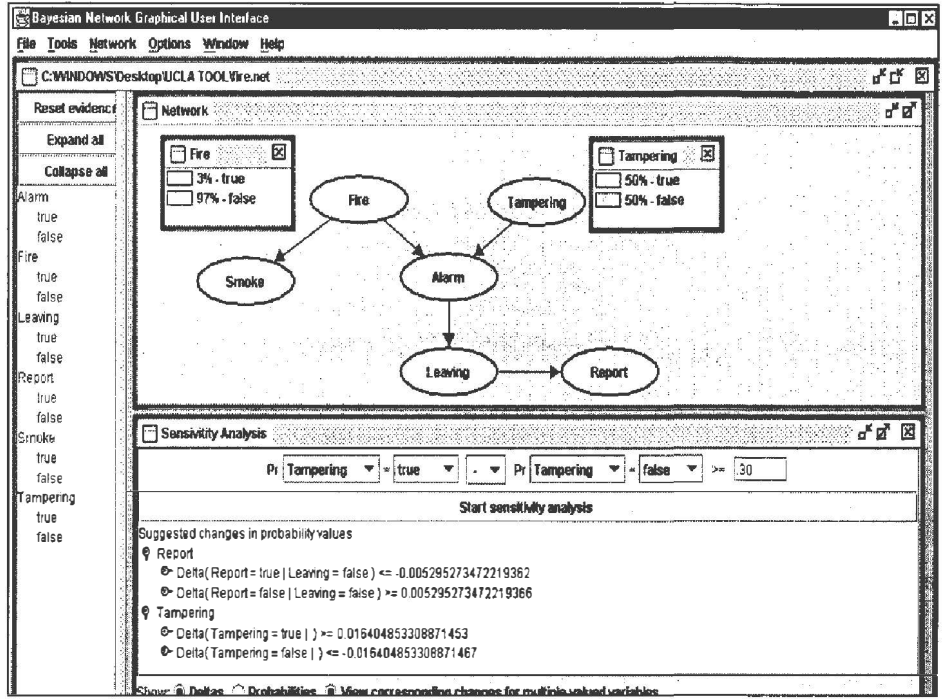

Figure 1: A sensitivity analysis scenario.

*however, for the probability of tampering to be no less than .65 (that is, $Pr(tampering \mid \mathbf{e}) - Pr(\overline{tampering} \mid \mathbf{e}) \geq .30$). SAMIAM recommended two ways to achieve this:*[2]

1. *either increase the prior probability of tampering by .016 or more (from its current value of .02), or*

2. *decrease the probability of a false report, $Pr(report \mid \overline{leaving})$, by .005 or more (from its current value of .01).*

*Therefore, the distinctions between .02 and .036, and the one between .01 and .005, do really matter in this case as each induces an absolute change of .15 on the probabilistic query of interest.*

Later, we show an example where an infinitesimal change to a network parameter leads to a change of .5 to a corresponding query. We also show examples in which the *relative* change in the probability of a query is larger than the corresponding relative change in a network parameter. One wonders then whether there is a different method for measuring probabilistic change (other than absolute or relative), which allows one to non-trivially bound the change in a probabilistic query in terms of the corresponding change in a network parameter.

---
[2]Implicit in the recommendation of SAMIAM is that the parameters of variables *fire*, *smoke*, *leaving*, and *alarm* are irrelevant to enforcing the given constraint.

To answer these and similar questions, we conduct in this paper an analytic study of the derivative $\partial Pr(y \mid \mathbf{e})/\partial \tau_{x\mid\mathbf{u}}$, where $\tau_{x\mid\mathbf{u}}$ is a meta parameter that allows us to simultaneously change co-varying parameters such as $\theta_{x\mid\mathbf{u}}$ and $\theta_{\overline{x}\mid\mathbf{u}}$. Our study leads us to three main results:

- Theorem 3.1, which provides a bound on the derivative $\partial Pr(y \mid \mathbf{e})/\partial \tau_{x\mid\mathbf{u}}$ in terms of $Pr(y \mid \mathbf{e})$ and $Pr(x \mid \mathbf{u})$ only.

- Theorem 3.2, which proves a bound on the sensitivity of queries to infinitesimal changes in network parameters.

- Theorem 3.3, which proves a bound on the sensitivity of queries to arbitrary changes in network parameters.

The last theorem shows that the amount of change in a probabilistic query can be bounded in terms of the amount of change in a network parameter, as long as change is understood to be the *relative change in odds*. This result has a number of practical implications. First, it can relief experts from having to be too precise when specifying certain parameters subjectively. Next, it can be important for approximate inference algorithms that pre-process network parameters to eliminate small distinctions in such parameters (in order to increase the efficiency of inference [10]). Finally, it can be used to show that automated reasoning systems based on belief networks are robust and, hence, suitable for real-world applications [11].

This paper is structured as follows. We present in Section 2 a technique for automatically identifying minimal parameter changes that are needed to ensure certain constraints on probabilistic queries. We then study analytically in Section 3 the impact of parameter changes on query changes, providing a series of results in this direction. Section 4 is then dedicated to exploring the implications of our results in Section 3; in particular, providing an analytic explanation of why certain parameter changes don't matter. We finally close in Section 5 with some concluding remarks. Proof sketches of theorems are given in Appendix A, while full proofs can be found in [2].

## 2 The tuning of network parameters

We report in this section on a tool that we have been developing, called SAMIAM for fine tuning network parameters [7]. Given a belief network, evidence $\mathbf{e}$ and two events $y$ and $z$, where $Y, Z \notin \mathbf{E}$, our tool can efficiently identify parameter changes needed to enforce the following types of constraints:

DIFFERENCE: $Pr(y \mid \mathbf{e}) - Pr(z \mid \mathbf{e}) \geq \epsilon$.

RATIO: $Pr(y \mid \mathbf{e})/Pr(z \mid \mathbf{e}) \geq \epsilon$.

We discuss next how one would enforce the DIFFERENCE constraint, which is very similar to enforcing the RATIO constraint.

When considering the parameters of variables $X$, we assume that it only has two values $x$ and $\overline{x}$ and, hence, two parameters $\theta_{x|\mathbf{u}}$ and $\theta_{\overline{x}|\mathbf{u}}$ for each parent instantiation $\mathbf{u}$. Moreover, we assume that for each variable $X$ and parent instantiation $\mathbf{u}$ we have a meta parameter $\tau_{x|\mathbf{u}}$, such that $\theta_{x|\mathbf{u}} = \tau_{x|\mathbf{u}}$ and $\theta_{\overline{x}|\mathbf{u}} = 1 - \tau_{x|\mathbf{u}}$. Therefore, our goal is then to determine the amount of change to the meta parameter $\tau_{x|\mathbf{u}}$ which would lead to a simultaneous change in both of $\theta_{x|\mathbf{u}}$ and $\theta_{\overline{x}|\mathbf{u}}$.[3] Our results can be easily extended to multivalued variables, as long as we assume a model for changing co-varying parameters when one of them changes [3, 8]. But we leave that extension out to simplify the discussion.

To enforce the DIFFERENCE constraint, we observe first that the probability of an instantiation $\mathbf{e}$, $Pr(\mathbf{e})$, is a linear function in any network parameter $\theta_{x|\mathbf{u}}$ [12, 1]. In fact, the probability is also linear in any meta parameter $\tau_{x|\mathbf{u}}$ and we have [2]:

$$\frac{\partial Pr(\mathbf{e})}{\partial \tau_{x|\mathbf{u}}} = \frac{Pr(\mathbf{e}, x, \mathbf{u})}{\theta_{x|\mathbf{u}}} - \frac{Pr(\mathbf{e}, \overline{x}, \mathbf{u})}{\theta_{\overline{x}|\mathbf{u}}}, \quad (1)$$

when $\theta_{x|\mathbf{u}} \neq 0$ and $\theta_{\overline{x}|\mathbf{u}} \neq 0$.[4] We will designate the above derivative by the constant $\alpha_\mathbf{e}$. Similarly, we will designate the derivatives $\partial Pr(y, \mathbf{e})/\partial \tau_{x|\mathbf{u}}$ and $\partial Pr(z, \mathbf{e})/\partial \tau_{x|\mathbf{u}}$ by the constants $\alpha_{y,\mathbf{e}}$ and $\alpha_{z,\mathbf{e}}$, respectively:

$$\frac{\partial Pr(y, \mathbf{e})}{\partial \tau_{x|\mathbf{u}}} = \frac{Pr(y, \mathbf{e}, x, \mathbf{u})}{\theta_{x|\mathbf{u}}} - \frac{Pr(y, \mathbf{e}, \overline{x}, \mathbf{u})}{\theta_{\overline{x}|\mathbf{u}}}, \quad (2)$$

$$\frac{\partial Pr(z, \mathbf{e})}{\partial \tau_{x|\mathbf{u}}} = \frac{Pr(z, \mathbf{e}, x, \mathbf{u})}{\theta_{x|\mathbf{u}}} - \frac{Pr(z, \mathbf{e}, \overline{x}, \mathbf{u})}{\theta_{\overline{x}|\mathbf{u}}}. \quad (3)$$

Now, to ensure that $Pr(y \mid \mathbf{e}) - Pr(z \mid \mathbf{e}) \geq \epsilon$, it suffices to ensure that $Pr(y, \mathbf{e}) - Pr(z, \mathbf{e}) \geq \epsilon Pr(\mathbf{e})$. Suppose that the previous constraint does not hold, and we wish to establish it by applying a change of $\delta$ to the meta parameter $\tau_{x|\mathbf{u}}$. Such a change leads to a change of $\alpha_\mathbf{e}\delta$ in $Pr(\mathbf{e})$. It also changes $Pr(y, \mathbf{e})$ and $Pr(z, \mathbf{e})$ by $\alpha_{y,\mathbf{e}}\delta$ and $\alpha_{z,\mathbf{e}}\delta$, respectively. Hence, to enforce the constraint, we need to solve for $\delta$ in the following inequality:

$$[Pr(y, \mathbf{e}) + \alpha_{y,\mathbf{e}}\delta] - [Pr(z, \mathbf{e}) + \alpha_{z,\mathbf{e}}\delta] \geq \epsilon[Pr(\mathbf{e}) + \alpha_\mathbf{e}\delta].$$

Rearranging the terms, we get:

$$Pr(y, \mathbf{e}) - Pr(z, \mathbf{e}) - \epsilon Pr(\mathbf{e}) \geq$$
$$\delta[-\alpha_{y,\mathbf{e}} + \alpha_{z,\mathbf{e}} + \epsilon\alpha_\mathbf{e}]. \quad (4)$$

Given Equations 1–4, we can then easily solve for the amount of change needed, $\delta$, once we know the following probabilities $Pr(y, \mathbf{e})$, $Pr(z, \mathbf{e})$, $Pr(\mathbf{e})$, $Pr(\mathbf{e}, x, \mathbf{u})$, $Pr(\mathbf{e}, \overline{x}, \mathbf{u})$, $Pr(y, \mathbf{e}, x, \mathbf{u})$, $Pr(y, \mathbf{e}, \overline{x}, \mathbf{u})$, $Pr(z, \mathbf{e}, x, \mathbf{u})$, and $Pr(z, \mathbf{e}, \overline{x}, \mathbf{u})$.

The question now is how to compute these probabilities, efficiently, and for all meta parameters $\tau_{x|\mathbf{u}}$ as there may be more than one possible parameter change that would enforce the given constraint; we need to identify all such parameters.

Interestingly enough, if we have an algorithm that can compute $Pr(\mathbf{i}, x, \mathbf{u})$, for a given instantiation $\mathbf{i}$ and for all family instantiations $x, \mathbf{u}$, then that algorithm can be used to evaluate Inequality 4 for every meta parameter $\tau_{x|\mathbf{u}}$. All we have to do is run the algorithm three times, once with $\mathbf{i} = \mathbf{e}$ and then again with $\mathbf{i} = y, \mathbf{e}$ and finally with $\mathbf{i} = z, \mathbf{e}$. Both the jointree algorithm [5] and the differential approach in [3] have the previous ability and can be used for this purpose.

---

[3]It is not meaningful to change one of the parameters in $\theta_{x|\mathbf{u}}, \theta_{\overline{x}|\mathbf{u}}$ without changing the other since $\theta_{x|\mathbf{u}} + \theta_{\overline{x}|\mathbf{u}} = 1$.

[4]If either of the previous parameters is zero, we can use the differential approach in [3] to compute the derivative directly [2].

We present now another example to illustrate how the results above are used in practice.

**Example 2.1** *Consider again the network in Figure 1. Here, we set the evidence such that we have smoke, but no report of people evacuating the building: $\mathbf{e} = smoke, \overline{report}$. We then got the posteriors $Pr(fire \mid \mathbf{e}) = .25$ and $Pr(tampering \mid \mathbf{e}) = .02$. We thought in this case that the posterior on fire should be no less than .50 and asked SAMIAM to recommend the necessary changes to enforce this constraint. There were five recommendations in this case, three of which could be ruled out based on qualitative considerations:*

1. *increase the prior on fire to $\geq .03$ (from .01);*

2. *increase the prior on tampering to $\geq .80$ (from .02);*

3. *decrease $Pr(smoke \mid \overline{fire})$ to $\leq .003$ (from .01);*

4. *increase $Pr(leaving \mid \overline{alarm})$ to $\geq .923$ (from .001);*

5. *increase $Pr(report \mid \overline{leaving})$ to $\geq .776$ (from .01).*

*Clearly, the only sensible changes here are either to increase the prior on fire, or to decrease the probability of having smoke without a fire.*

This example and other similar ones suggest that identifying such parameter changes and their magnitude is inevitable for developing faithful belief networks, yet is not trivial to accomplish by a visual inspection of the belief network and, hence, need to be facilitated by sensitivity analysis tools. Moreover, the examples illustrate the need to develop more analytic tools to understand and explain the sensitivity of queries to certain parameter changes. There is also a need to reconcile the sensitivities exhibited by our examples with previous experimental studies demonstrating the robustness of probabilistic queries against small parameter changes in certain application areas, such as diagnosis [11]. We address these particular questions in the next two sections.

## 3 The sensitivity of probabilistic queries to parameters changes

Our starting point in understanding the sensitivity of a query $Pr(y \mid \mathbf{e})$ to changes in a meta parameter $\tau_{x|\mathbf{u}}$ is to analyze the derivative $\partial Pr(y \mid \mathbf{e})/\partial \tau_{x|\mathbf{u}}$. In our analysis, we assume that $X$ is binary, but $Y$ and all other variables in the network can be multivalued. The following theorem provides a simple bound on this derivative, in terms of $Pr(y \mid \mathbf{e})$ and $Pr(x \mid \mathbf{u})$ only. We then use this simple bound to study the effect of changes to meta parameters on probabilistic queries.

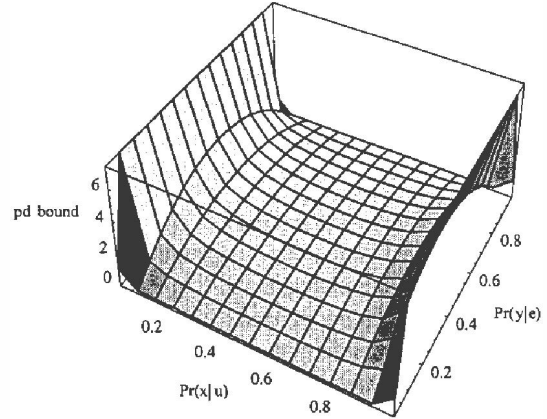

Figure 2: The plot of the upper bound on the partial derivative $\partial Pr(y \mid \mathbf{e})/\partial \tau_{x|\mathbf{u}}$, as given in Theorem 3.1, against $Pr(x \mid \mathbf{u})$ and $Pr(y \mid \mathbf{e})$.

**Theorem 3.1** *If $X$ is a binary variable in a belief network, then:*[5]

$$\left| \frac{\partial Pr(y \mid \mathbf{e})}{\partial \tau_{x|\mathbf{u}}} \right| \leq \frac{Pr(y \mid \mathbf{e})(1 - Pr(y \mid \mathbf{e}))}{Pr(x \mid \mathbf{u})(1 - Pr(x \mid \mathbf{u}))}.$$

We show later an example for which the derivative assumes the above bound exactly.[6]

The plot of this bound against $Pr(x \mid \mathbf{u})$ and $Pr(y \mid \mathbf{e})$ is shown in Figure 2. A number of observations are in order about this plot:

- For extreme values of $Pr(x \mid \mathbf{u})$, the bound approaches infinity, and thus a small absolute change in the meta parameter $\tau_{x|\mathbf{u}}$ can have a big impact on the query $Pr(y \mid \mathbf{e})$.

- On the other hand, the bound approaches 0 for extreme values of the query $Pr(y \mid \mathbf{e})$. Therefore, a small absolute change in the meta parameter

---

[5]This theorem and all results that follow requires that $\tau_{x|\mathbf{u}} \neq 0$ and $\tau_{x|\mathbf{u}} \neq 1$, since we can only use the expression in Equation 1 under these conditions.

[6]Note that we have an exact closed form for the derivative $\partial Pr(y \mid \mathbf{e})/\partial \tau_{x|\mathbf{u}}$ [3, 4], but that form includes terms which are specific to the given belief network.

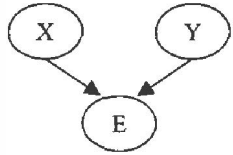

Figure 3: The network used in Example 3.1.

$\tau_{x|\mathbf{u}}$ will have a small effect on the absolute change in the query.

One of the implications of this result is that if we have a belief network where queries $Pr(y \mid \mathbf{e})$ have extreme values, then such networks will be robust against small changes in network parameters. This of course assumes that robustness is understood to be a small change in the absolute value of the given query. Interestingly enough, if $y$ is a disease which is diagnosed by finding $\mathbf{e}$—that is, the probability $Pr(y \mid \mathbf{e})$ is quite high—then it is not surprising that such queries would be robust against small perturbations to network parameters. This seems to explain some of the results in [11], where robustness have been confirmed for queries with $Pr(y \mid \mathbf{e}) \geq .90$.

Another implication of the above result is that one has to be careful when changing parameters that are extreme. Such parameters are potentially very influential and one must handle them with care.

Therefore, the worst situation from a robustness viewpoint materializes if one has extreme parameters with non-extreme queries. In such a case, the queries can be very sensitive to small variations in the parameters.

**Example 3.1** Consider the network structure in Figure 3. We have two binary nodes, $X$ and $Y$ with respective parameters $\theta_x, \theta_{\overline{x}}$ and $\theta_y, \theta_{\overline{y}}$. We assume that $E$ is a deterministic binary node where the value of $E$ is $e$ iff $X = Y$. This dictates the following CPT for $E$: $Pr(e \mid x, y) = 1$, $Pr(e \mid \overline{x}, \overline{y}) = 1$, $Pr(e \mid x, \overline{y}) = 0$ and $Pr(e \mid \overline{x}, y) = 0$. The conditional probability $Pr(y \mid \mathbf{e})$ can be expressed using the root parameters $\theta_x$ and $\theta_y$ as:

$$Pr(y \mid e) = \frac{\theta_x \theta_y}{\theta_x \theta_y + \theta_{\overline{x}} \theta_{\overline{y}}}.$$

The derivative of this probability with respect to the meta parameter $\tau_x$ is given by:[7]

$$\frac{\partial Pr(y \mid e)}{\partial \tau_x} = \frac{(\theta_x \theta_y + \theta_{\overline{x}} \theta_{\overline{y}})\theta_y - \theta_x \theta_y (\theta_y - \theta_{\overline{y}})}{(\theta_x \theta_y + \theta_{\overline{x}} \theta_{\overline{y}})^2}$$
$$= \frac{\theta_y \theta_{\overline{y}}}{(\theta_x \theta_y + \theta_{\overline{x}} \theta_{\overline{y}})^2}.$$

---
[7] Note that $\partial \theta_x / \partial \tau_x = 1$ and $\partial \theta_{\overline{x}} / \partial \tau_x = -1$.

This is equal to the upper bound given in Theorem 3.1:

$$\frac{Pr(y \mid e)(1 - Pr(y \mid e))}{Pr(x)(1 - Pr(x))} = \frac{(\theta_x \theta_y)(\theta_{\overline{x}} \theta_{\overline{y}})}{\theta_x \theta_{\overline{x}}(\theta_x \theta_y + \theta_{\overline{x}} \theta_{\overline{y}})^2}$$
$$= \frac{\theta_y \theta_{\overline{y}}}{(\theta_x \theta_y + \theta_{\overline{x}} \theta_{\overline{y}})^2}.$$

Now, if we set $\theta_x = \theta_{\overline{y}}$, the derivative becomes:

$$\frac{\partial Pr(y \mid e)}{\partial \tau_x} = \frac{1}{4\theta_x \theta_{\overline{x}}},$$

and as $\theta_x$ (or $\theta_{\overline{x}}$) approaches 0, the derivative approaches infinity. Finally, if we set $\theta_x = \theta_{\overline{y}} = \epsilon$, we have $Pr(y|e) = .5$, but if we keep $\theta_y$ and $\theta_{\overline{y}}$ constant and change $\tau_x$ to 0, we get the new result $Pr(y|e) = 0$.

Example 3.1 then illustrates three points. First, it shows that the bound in Theorem 3.1 is tight in the above sense. Second, it gives an example network which the derivative $\partial Pr(y \mid \mathbf{e})/\partial \tau_{x|\mathbf{u}}$ tends to infinity. Finally, it shows that an infinitesimal absolute change in a parameter can induce a non-infinitesimal absolute change in some query. The following theorem, however, shows that this is not possible if we consider a *relative* notion of change.

**Theorem 3.2** *Assume that $\tau_{x|\mathbf{u}} \leq .5$ without loss of generality.*[8] *Suppose that $\Delta \tau_{x|\mathbf{u}}$ is an infinitesimal change applied to the meta parameter $\tau_{x|\mathbf{u}}$, leading to a change of $\Delta Pr(y \mid \mathbf{e})$ to the query $Pr(y \mid \mathbf{e})$. We then have:*

$$\left| \frac{\Delta Pr(y \mid \mathbf{e})}{Pr(y \mid \mathbf{e})} \right| \leq 2 \left| \frac{\Delta \tau_{x|\mathbf{u}}}{\tau_{x|\mathbf{u}}} \right|.$$

For a function $f(x)$, the quantity:

$$\lim_{(x-x_0) \to 0} \frac{(f(x) - f(x_0))/f(x_0)}{(x - x_0)/x_0},$$

is typically known as the *sensitivity* of $f$ to $x$ at $x_0$. Theorem 3.2 is then showing that the sensitivity of $Pr(y \mid \mathbf{e})$ to $\tau_{x|\mathbf{u}}$ is bounded.

As an example application of Theorem 3.2, consider Example 3.1 again. The change of $\tau_x$ from $\epsilon$ to 0 amounts to a relative change $|-\epsilon/\epsilon| = 1$. The corresponding change of $Pr(y \mid e)$ from .5 to 0 amounts to a relative change of $|-.5/.5| = 1$.[9] Hence, the relative change in the query is not as dramatic from this viewpoint.

---
[8] For a binary variable $X$, if $\tau_{x|\mathbf{u}} > .5$, we can instead choose the meta parameter $\tau_{\overline{x}|\mathbf{u}}$ without loss of generality.

[9] If we consider the meta parameter $\tau_{\overline{x}} = 1 - \epsilon$ instead, the relative change in $\tau_{\overline{x}}$ will then amount to $\epsilon/(1-\epsilon)$. But Theorem 3.2 will not be applicable in this case (assuming that $\epsilon$ is close to 0) since the theorem requires that the chosen meta parameter be $\leq .5$.

The relative change in $Pr(y \mid e)$ may be greater than double the relative change in $\tau_{x|u}$ for non-infinitesimal changes because the derivative $\partial Pr(y \mid e)/\partial \tau_{x|u}$ depends on the value of $\tau_{x|u}$ [3, 7]. Going back to Example 3.1, if we set $\theta_x = .5$ and $\theta_y = .01$, we obtain the result $Pr(y \mid e) = .01$. If we now increase $\tau_x$ to .6, a relative change of 20%, we get the new result $Pr(y \mid e) = 0.0149$, a relative change of 49%, which is more than double of the relative change in $\tau_x$.

The question now is: Suppose that we change a meta parameter $\tau_{x|u}$ by an arbitrary amount (not an infinitesimal amount), what can we say about the corresponding change in the query $Pr(y \mid e)$? We have the following result.

**Theorem 3.3** *Let $O(x \mid u)$ denote the odds of $x$ given $u$: $O(x \mid u) = Pr(x \mid u)/(1 - Pr(x \mid u))$, and let $O(y \mid e)$ denote the odds of $y$ given $e$: $O(y \mid e) = Pr(y \mid e)/(1 - Pr(y \mid e))$. Let $O'(x \mid u)$ and $O'(y \mid e)$ denote these odds after having applied an arbitrary change to the meta parameter $\tau_{x|u}$ where $X$ is a binary variable in a belief network. If the change is positive, then:*

$$\frac{O(x \mid u)}{O'(x \mid u)} \leq \frac{O'(y \mid e)}{O(y \mid e)} \leq \frac{O'(x \mid u)}{O(x \mid u)},$$

*and if it is negative, then:*

$$\frac{O'(x \mid u)}{O(x \mid u)} \leq \frac{O'(y \mid e)}{O(y \mid e)} \leq \frac{O(x \mid u)}{O'(x \mid u)}.$$

*Combining both results, we have:*

$$|\ln(O'(y \mid e)) - \ln(O(y \mid e))| \leq \\ |\ln(O'(x \mid u)) - \ln(O(x \mid u))|.$$

Theorem 3.3 means that the relative change in the odds of $y$ given $e$ is bounded by the relative change in the odds of $x$ given $u$. Note that the result makes no assumptions whatsoever about the structure of the given belief network.

An interesting special case of the above result is when $X$ is a root node and $X = Y$. From basic probability theory, we have:

$$O(x \mid e) = O(x) \frac{Pr(e \mid x)}{Pr(e \mid \overline{x})}.$$

As the ratio $Pr(e \mid x)/Pr(e \mid \overline{x})$ is independent of $Pr(x)$, the ratio $O(x \mid e)/O(x)$ is also independent of this prior. Therefore, we can conclude that:

$$\frac{O'(x \mid e)}{O(x \mid e)} = \frac{O'(x)}{O(x)}. \quad (5)$$

This means we can find the exact amount of change needed for a meta parameter $\tau_x$ in order to induce a particular change on the query $Pr(x \mid e)$. There is no need to use the more expensive technique of Section 2 in this case.

**Example 3.2** *Consider the network in Figure 1. Suppose that $e = \text{report}, \overline{\text{smoke}}$. Currently, $Pr(\text{tampering}) = .02$ and $Pr(\text{tampering} \mid e) = .50$. We wish to increase the conditional probability to .65. We can compute the new prior probability $Pr'(\text{tampering})$ using Equation 5:*

$$\frac{.65/.35}{.50/.50} = \frac{Pr'(\text{tampering})/(1 - Pr'(\text{tampering}))}{.98/.02},$$

*giving us $Pr'(\text{tampering}) = .036$, which is equal to the result we obtained using SAMIAM in Example 1.1. Both the changes to $Pr(\text{tampering})$ and $Pr(\text{tampering} \mid e)$ bring a log-odds difference of .616.*

Theorem 3.3 has a number of implications. First, given a particular query $Pr(y \mid e)$ and a meta parameter $\tau_{x|u}$, it can be used to bound the effect that a change in $\tau_{x|u}$ will have on the query $Pr(y \mid e)$. Going back to Example 3.2, we may wish to know what is the impact on other conditional probabilities if we apply the change making $Pr'(\text{tampering}) = .036$. The log-odds changes for all conditional probabilities in the network will be bounded by .616. For example, currently $Pr(\text{fire} \mid e) = .029$. Using Theorem 3.3, we can find the range of the new conditional probability value $Pr'(\text{fire} \mid e)$:

$$\left| \ln\left(\frac{Pr'(\text{fire} \mid e)}{1 - Pr'(\text{fire} \mid e)}\right) - \ln\left(\frac{.029}{.971}\right) \right| \leq .616,$$

giving us the range $.016 \leq Pr'(\text{fire} \mid e) \leq .053$. The exact value of $Pr'(\text{fire} \mid e)$, obtained by inference, is .021, which is within the computed bounds.

Second, Theorem 3.3 can be used to efficiently approximate solutions to the DIFFERENCE and RATIO problems we discussed in Section 2. That is, given a desirable change in the value of query $Pr(y \mid e)$, we can use Theorem 3.3 to immediately compute a lower bound on the minimum change to meta parameter $\tau_{x|u}$ needed to induce the change. This method can be applied in constant time and can serve as a preliminary recommendation, as the method proposed in Section 2 is much more expensive computationally.

Third, suppose that SAMIAM was used to recommend parameter changes that would induce a desirable change on a given query. Suppose further that SAMIAM returned a number of such changes, each of which is capable of inducing the necessary change. The question is: which one of these changes should we adopt? The main principle applied in these situations is to adopt a "minimal" change. But what is minimal in this case? As Theorem 3.3 reveals, a notion of minimality which is based on the amount of absolute change can be very misleading. Instead, it suggests that one adopts the change that minimizes the relative change in the odds, as other queries can be shown to be robust against such a change in a precise sense.

Finally, the result can be used to obtain a better intuitive understanding of parameter changes that do or do not matter, a topic which we will discuss in Section 4.

## 4 Changes that (don't) matter

We now return to a central question: When do changes in network parameters matter and when do they not matter? As we mentioned earlier, there have been experimental studies investigating the robustness of belief networks against parameter changes. But we have also shown very simple and intuitive examples where networks can be very sensitive to small parameter changes. This calls for a better understanding of the effect of parameter changes on queries, so one can intuitively sort out situations in which such changes do or do not matter. Our goal in this section is to further develop such an understanding by looking more closely into some of the implications of Theorem 3.3.

First, we have to settle the issue of "What does it mean for a parameter change to matter?" One can think of at least three definitions. First, the absolute change in the probability $Pr(y \mid e)$ is small. Second, the relative change in the probability $Pr(y \mid e)$ is small. Third, relative change in the odds of $y$ given $e$ is small. The first notion is the one most prevalent in the literature, so we shall adopt it in the rest of this section.

Suppose we have a belief network for a diagnostic application and suppose we are concerned about the robustness of the query $Pr(y \mid e)$ with respect to changes in network parameters. In this application, $y$ is a particular disease and $e$ is a particular finding which predicts the disease, with $Pr(y \mid e) = .90$. Let us define robustness in this case to be an absolute change of no more than .05 to the given query. Now, let $X$ be a binary variable in the network and let us ask: What kind of changes to the parameters on $X$ are guaranteed to keep the query within the desirable range? We can use Theorem 3.3 easily to answer this question. First, if we are changing a parameter by $\delta$, and if we want the value of the query to remain $\leq .95$, we must ensure that:

$$|\ln((p+\delta)/(1-p-\delta)) - \ln(p/(1-p))| \leq .7472,$$

where $.7472 = |\ln(.95/.05) - \ln(.90/.10)|$ and $p$ is the current value of the parameter.

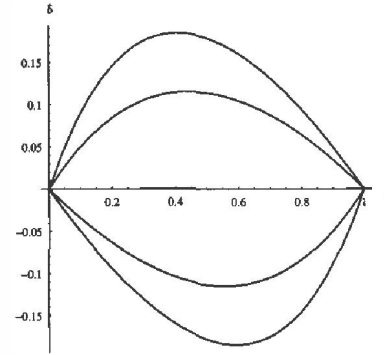

Figure 4: The amount of parameter change $\delta$ that would guarantee the query $Pr(y \mid e) = .90$ to stay within the interval $[.85, .95]$, as a function of the current parameter value $p$. The outer envelope guarantees the query to remain $\leq .95$, while the inner envelope guarantees the query to remain $\geq .85$.

Similarly, if we want to ensure that the query remains $\geq .85$, we want to ensure that:

$$|\ln((p+\delta)/(1-p-\delta)) - \ln(p/(1-p))| \leq .4626,$$

where $.4626 = |\ln(.85/.15) - \ln(.90/.10)|$.

Figure 4 plots the permissible change $\delta$ as a function of $p$, the current value of the parameter. The main thing to observe here is that the amount of permissible change depends on the current value of $p$, with smaller changes allowed for extreme values of $p$. It is also interesting to note that it is easier to guarantee the query to stay $\leq .95$ than to guarantee that it stays $\geq .85$. Therefore, it is more likely for a parameter change to reduce the value of a query which is close to 1 (and to increase the value of a query which is close to 0). Finally, if we are increasing the parameter, then a parameter value close to .4 would allow the biggest change. But if we are decreasing the parameter, then a value close to .6 will allow the biggest change.

Now let us repeat the same exercise but assuming that the initial value of the query is $Pr(y \mid e) = .60$, yet insisting on the same measure of robustness. Figure 5 plots the permissible changes $\delta$ as a function of $p$, the current value of the parameter. Again, the amount of permissible change becomes smaller as the probability $p$ approaches 0 or 1. The other main point to emphasize is that the permissible changes are now much smaller than in the previous example, since the initial value of the query is not as extreme. Therefore, this query is much less robust than the previous one.

More generally, Figure 6 plots the log-odd difference, $|\ln(O'(x \mid u)) - \ln(O(x \mid u))|$ against $Pr(x \mid u) = p$ and $Pr'(x \mid u) = p + \delta$. Again, the plot explains

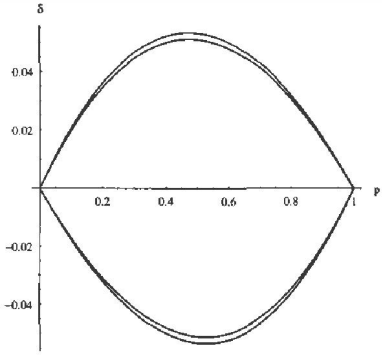

Figure 5: The amount of parameter change $\delta$ that would guarantee the query $Pr(y \mid \mathbf{e}) = .60$ to stay within the interval $[.55, .65]$, as a function of the current parameter value $p$. The outer envelope guarantees the query to remain $\leq .65$, while the inner envelope guarantees the query to stay in $\geq .55$.

analytically why we can afford more absolute changes to non-extreme probabilities [11, 10].

We close this section by emphasizing that the above figures identify parameter changes that guarantee to keep queries within certain ranges. However, if the belief network has specific properties, such as a specific topology, then it is possible for the query to be robust against parameter changes that are outside the identified bounds.

## 5 Conclusion

We presented an efficient technique for fine-tuning the parameters of a belief network. The technique suggests minimal changes to network parameters which ensure that certain constraints are enforced on probabilistic queries. Based on this technique, we have experimented with some belief networks, only to find out that these networks are more sensitive to parameter changes than previous experimental studies seem to suggest. This observation lead us to an analytic study on the effect of parameter changes, with the aim of characterizing situations under which parameter changes do or do not matter. We have reported on a number of results in this direction. Our central result shows that belief networks are robust in a very specific sense: the relative change in query odds is bounded by the relative change in the parameter odds. A closer look at this result, its meaning, and its implications provide interesting characterizations of parameter changes that do or do not matter, and explains analytically some of the previous experimental results and observations on this matter.

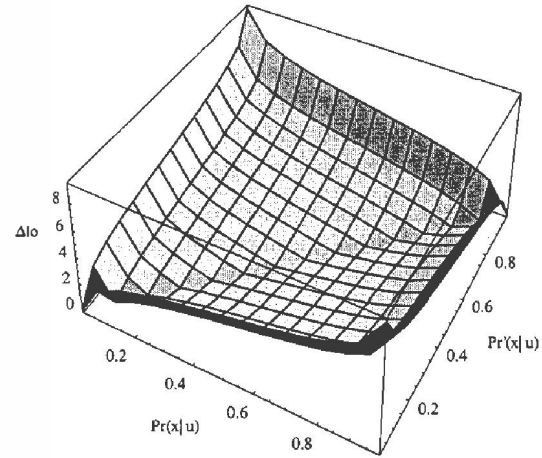

Figure 6: The plot of the log-odd difference, $\Delta lo = |\ln(O'(x \mid \mathbf{u})) - \ln(O(x \mid \mathbf{u}))|$, against $Pr(x \mid \mathbf{u})$ and $Pr'(x \mid \mathbf{u})$.

## A Proof sketches

Detailed proofs can be found in [2].

**Theorem 3.1**

From [3], the derivative $\partial Pr(y \mid \mathbf{e}) / \partial \theta_{x \mid \mathbf{u}}$ is equal to:

$$\frac{\partial Pr(y \mid \mathbf{e})}{\partial \theta_{x \mid \mathbf{u}}} = \frac{Pr(y, x, \mathbf{u} \mid \mathbf{e}) - Pr(y \mid \mathbf{e}) Pr(x, \mathbf{u} \mid \mathbf{e})}{Pr(x \mid \mathbf{u})}.$$

Since, we have:

$$\frac{\partial Pr(y \mid \mathbf{e})}{\partial \tau_{x \mid \mathbf{u}}} = \frac{\partial Pr(y \mid \mathbf{e})}{\partial \theta_{x \mid \mathbf{u}}} - \frac{\partial Pr(y \mid \mathbf{e})}{\partial \theta_{\overline{x} \mid \mathbf{u}}},$$

we can verify that:

$$\begin{aligned}&\frac{\partial Pr(y \mid \mathbf{e})}{\partial \tau_{x \mid \mathbf{u}}} \\ &= \frac{Pr(y, x, \mathbf{u} \mid \mathbf{e}) - Pr(y \mid \mathbf{e}) Pr(x, \mathbf{u} \mid \mathbf{e})}{Pr(x \mid \mathbf{u})(1 - Pr(x \mid \mathbf{u}))} - \\ &\quad \frac{Pr(x \mid \mathbf{u})(Pr(y, \mathbf{u} \mid \mathbf{e}) - Pr(y \mid \mathbf{e}) Pr(\mathbf{u} \mid \mathbf{e}))}{Pr(x \mid \mathbf{u})(1 - Pr(x \mid \mathbf{u}))}.\end{aligned}$$

In order to find an upper bound on the derivative, we would like to bound the term $Pr(y, x, \mathbf{u} \mid \mathbf{e}) - Pr(y \mid \mathbf{e}) Pr(x, \mathbf{u} \mid \mathbf{e})$. We have:

$$\begin{aligned} & Pr(y, x, \mathbf{u} \mid \mathbf{e}) - Pr(y \mid \mathbf{e}) Pr(x, \mathbf{u} \mid \mathbf{e}) \\ \leq\ & Pr(y, x, \mathbf{u} \mid \mathbf{e}) - Pr(y \mid \mathbf{e}) Pr(y, x, \mathbf{u} \mid \mathbf{e}) \\ =\ & Pr(y, x, \mathbf{u} \mid \mathbf{e})(1 - Pr(y \mid \mathbf{e})) \\ \leq\ & Pr(y, \mathbf{u} \mid \mathbf{e})(1 - Pr(y \mid \mathbf{e})). \end{aligned}$$

Therefore, the upper bound on the derivative can be verified as:

$$\begin{aligned}
\frac{\partial Pr(y \mid \mathbf{e})}{\partial \tau_{x|\mathbf{u}}} & \leq \frac{Pr(y, \mathbf{u} \mid \mathbf{e})(1 - Pr(y \mid \mathbf{e}))}{Pr(x \mid \mathbf{u})(1 - Pr(x \mid \mathbf{u}))} - \\
& \quad \frac{Pr(x \mid \mathbf{u})(Pr(y, \mathbf{u} \mid \mathbf{e}) - Pr(y \mid \mathbf{e})Pr(\mathbf{u} \mid \mathbf{e}))}{Pr(x \mid \mathbf{u})(1 - Pr(x \mid \mathbf{u}))} \\
& = \frac{Pr(\overline{y} \mid \mathbf{e})Pr(y, \mathbf{u} \mid \mathbf{e})}{Pr(x \mid \mathbf{u})} + \frac{Pr(y \mid \mathbf{e})Pr(\overline{y}, \mathbf{u} \mid \mathbf{e})}{1 - Pr(x \mid \mathbf{u})}.
\end{aligned}$$

Since $Pr(y, \mathbf{u} \mid \mathbf{e}) \leq Pr(y \mid \mathbf{e})$ and $Pr(\overline{y}, \mathbf{u} \mid \mathbf{e}) \leq Pr(\overline{y} \mid \mathbf{e})$, the upper bound on the derivative is:

$$\begin{aligned}
\frac{\partial Pr(y \mid \mathbf{e})}{\partial \tau_{x|\mathbf{u}}} & \leq \frac{Pr(\overline{y} \mid \mathbf{e})Pr(y \mid \mathbf{e})}{Pr(x \mid \mathbf{u})} + \frac{Pr(y \mid \mathbf{e})Pr(\overline{y} \mid \mathbf{e})}{1 - Pr(x \mid \mathbf{u})} \\
& = \frac{Pr(y \mid \mathbf{e})(1 - Pr(y \mid \mathbf{e}))}{Pr(x \mid \mathbf{u})(1 - Pr(x \mid \mathbf{u}))}.
\end{aligned}$$

Similarly, we can prove the lower bound on the derivative as:

$$\frac{\partial Pr(y \mid \mathbf{e})}{\partial \tau_{x|\mathbf{u}}} \geq -\frac{Pr(y \mid \mathbf{e})(1 - Pr(y \mid \mathbf{e}))}{Pr(x \mid \mathbf{u})(1 - Pr(x \mid \mathbf{u}))}.$$

**Theorem 3.2**

Because $\Delta \tau_{x|\mathbf{u}}$ is infinitesimal, from Theorem 3.1:

$$\begin{aligned}
\left| \frac{\Delta Pr(y \mid \mathbf{e})}{\Delta \tau_{x|\mathbf{u}}} \right| & \simeq \left| \frac{\partial Pr(y \mid \mathbf{e})}{\partial \tau_{x|\mathbf{u}}} \right| \\
& \leq \frac{Pr(y \mid \mathbf{e})(1 - Pr(y \mid \mathbf{e}))}{Pr(x \mid \mathbf{u})(1 - Pr(x \mid \mathbf{u}))}.
\end{aligned}$$

Arranging the terms, we have:

$$\begin{aligned}
\left| \frac{\Delta Pr(y \mid \mathbf{e})}{Pr(y \mid \mathbf{e})} \right| & \leq \frac{1 - Pr(y \mid \mathbf{e})}{1 - Pr(x \mid \mathbf{u})} \left| \frac{\Delta \tau_{x|\mathbf{u}}}{\tau_{x|\mathbf{u}}} \right| \\
& \leq \frac{1}{.5} \left| \frac{\Delta \tau_{x|\mathbf{u}}}{\tau_{x|\mathbf{u}}} \right| = 2 \left| \frac{\Delta \tau_{x|\mathbf{u}}}{\tau_{x|\mathbf{u}}} \right|,
\end{aligned}$$

since $Pr(x \mid \mathbf{u}) = \tau_{x|\mathbf{u}} \leq .5$.

**Theorem 3.3**

We obtain this result by integrating the bound in Theorem 3.1. In particular, if we change $\tau_{x|\mathbf{u}}$ to $\tau'_{x|\mathbf{u}} > \tau_{x|\mathbf{u}}$, and consequently $Pr(y \mid \mathbf{e})$ changes to $Pr'(y \mid \mathbf{e})$, we can separate the variables in the upper bound on the derivative in Theorem 3.1, and integrate over the intervals, and yield:

$$\int_{Pr(y|\mathbf{e})}^{Pr'(y|\mathbf{e})} \frac{dPr(y \mid \mathbf{e})}{Pr(y \mid \mathbf{e})(1 - Pr(y \mid \mathbf{e}))} \leq \int_{\tau_{x|\mathbf{u}}}^{\tau'_{x|\mathbf{u}}} \frac{d\tau_{x|\mathbf{u}}}{\tau_{x|\mathbf{u}}(1 - \tau_{x|\mathbf{u}})}.$$

This gives us the solution:

$$\begin{aligned}
& \ln(Pr'(y \mid \mathbf{e})) - \ln(Pr(y \mid \mathbf{e})) - \\
& \ln(1 - Pr'(y \mid \mathbf{e})) + \ln(1 - Pr(y \mid \mathbf{e})) \\
\leq \ & \ln(\tau'_{x|\mathbf{u}}) - \ln(\tau_{x|\mathbf{u}}) - \ln(1 - \tau'_{x|\mathbf{u}}) + \ln(1 - \tau_{x|\mathbf{u}}),
\end{aligned}$$

and after taking exponentials, we have:

$$\frac{Pr'(y \mid \mathbf{e})/(1 - Pr'(y \mid \mathbf{e}))}{Pr(y \mid \mathbf{e})/(1 - Pr(y \mid \mathbf{e}))} \leq \frac{\tau'_{x|\mathbf{u}}/(1 - \tau'_{x|\mathbf{u}})}{\tau_{x|\mathbf{u}}/(1 - \tau_{x|\mathbf{u}})},$$

which is equivalent to:

$$\frac{O'(y \mid \mathbf{e})}{O(y \mid \mathbf{e})} \leq \frac{O'(x)}{O(x)}.$$

The other parts of Theorem 3.3 can be proved similarly.

# References

[1] E. Castillo, J. M. Gutiérrez, and A. S. Hadi. Sensitivity analysis in discrete Bayesian networks. *IEEE Transactions on Systems, Man, and Cybernetics*, 27:412–423, 1997.

[2] Hei Chan and Adnan Darwiche. When do numbers really matter? Technical Report D-120, Computer Science Department, UCLA, Los Angeles, Ca 90095, 2001.

[3] Adnan Darwiche. A differential approach to inference in bayesian networks. In *Proceedings of the 16th Conference on Uncertainty in Artificial Intelligence (UAI)*, pages 123–132, 2000.

[4] Russell Greiner, Adam Grove and Dale Schuurmans. Learning Bayesian nets that perform well. In *Uncertainty in Artificial Intelligence*, 1997.

[5] F. V. Jensen, S.L. Lauritzen, and K.G. Olesen. Bayesian updating in recursive graphical models by local computation. *Computational Statistics Quarterly*, 4:269–282, 1990.

[6] Finn V. Jensen. *An Introduction to Bayesian Networks*. Springer Verlag, New York Inc., 1996.

[7] Finn V. Jensen. Gradient descent training of bayesian networks. In *In Proceedings of the Fifth European Conference on Symbolic and Quantitative Approaches to Reasoning with Uncertainty (ECSQARU)*, pages 5–9, 1999.

[8] Uffe Kjaerulff and Linda C. van der Gaag. Making sensitivity analysis computationally efficient. In *Proceedings of the 16th Conference on Uncertainty in Artificial Intelligence (UAI)*, 2000.

[9] Judea Pearl. *Probabilistic Reasoning in Intelligent Systems: Networks of Plausible Inference*. Morgan Kaufmann Publishers, Inc., San Mateo, California, 1988.

[10] David Poole. Context-specific approximation in probabilistic inference. In *Proceedings of the 14th Conference on Uncertainty in Artificial Intelligence (UAI)*, pages 447–454, 1998.

[11] Malcolm Pradhan, Max Henrion, Gregory Provan, Brendan Del Favero, and Kurt Huang. The sensitivity of belief networks to imprecise probabilities: an experimental investigation. *Artificial Intelligence*, 85:363–397, 1996.

[12] S. Russell, J. Binder, D. Koller, and K. Kanazawa. Local learning in probabilistic networks with hidden variables. In *Proceedings of the 11th Conference on Uncertainty in Artificial Intelligence (UAI)*, pages 1146–1152, 1995.